\documentclass[runningheads, envcountsame, a4paper]{llncs}

\usepackage{soul}
\usepackage{url}
\usepackage{graphicx}
\usepackage{enumitem}
\usepackage{algorithmic}
\usepackage{amsmath}
\usepackage{amsfonts}
\usepackage{makecell}
\usepackage{multirow} 
\usepackage{listings}
\usepackage{mathrsfs}
\usepackage{mathtools}
\usepackage{textcomp}
\usepackage{balance}
\usepackage{xcolor}
\usepackage[misc]{ifsym}

\newlength\savewidth
\newcommand\shline{\noalign{\global\savewidth\arrayrulewidth
                            \global\arrayrulewidth 1.5pt}%
                   \hline
                   \noalign{\global\arrayrulewidth\savewidth}}

\begin{document}
\title{Revisiting Convolutional Neural Networks for Citywide Crowd Flow Analytics}

\author{Yuxuan Liang\inst{1} \Letter, Kun Ouyang\inst{1}, Yiwei Wang\inst{1}, Ye Liu\inst{1}, Junbo Zhang\inst{2,3,4}, \\ Yu Zheng\inst{2,3,4}, David S. Rosenblum\inst{1}}
\institute{School of Computing, National University of Singapore, Singapore \and
JD Intelligent Cities Research \& JD Intelligent Cities Business Unit, Beijing, China \and Institute of Artificial Intelligence, Southwest Jiaotong University, China \and Xidian University, Xi’an, China
\email{\{yuxliang,msjunbozhang,msyuzheng\}@outlook.com \\
\{ouyangk,y-wang,liuye,david\}@comp.nus.edu.sg}}

\authorrunning{Liang et al.}
\titlerunning{Revisiting CNNs for Citywide Crowd Flow Analytics}

\maketitle            

\begin{abstract}
Citywide crowd flow analytics is of great importance to smart city efforts. It aims to model the crowd flow (e.g., inflow and outflow) of each region in a city based on historical observations. Nowadays, Convolutional Neural Networks (CNNs) have been widely adopted in raster-based crowd flow analytics by virtue of their capability in capturing spatial dependencies. After revisiting CNN-based methods for different analytics tasks, we expose two common critical drawbacks in the existing uses: 1) inefficiency in learning global spatial dependencies, and 2) overlooking latent region functions. To tackle these challenges, in this paper we present a novel framework entitled DeepLGR that can be easily generalized to address various citywide crowd flow analytics problems. This framework consists of three parts: 1) a local feature extraction module to learn representations for each region; 2) a global context module to extract global contextual priors and upsample them to generate the global features; and 3) a region-specific predictor based on tensor decomposition to provide customized predictions for each region, which is very parameter-efficient compared to previous methods. Extensive experiments on two typical crowd flow analytics tasks demonstrate the effectiveness, stability, and generality of our framework. 
\end{abstract}

\section{Introduction}\label{para:intro}
Citywide crowd flow analytics is very critical to smart city efforts around the world. A typical task is citywide crowd flow prediction \cite{zhang2016dnn,zhang2017deep,lin2019deepstn}, which aims to predict the traffic (e.g., inflows and outflows of every region) for the next time slot, given the historical traffic observations. It can help the governors conduct traffic control and avoid potential catastrophic stampede before a special event. Another important task is to infer the fine-grained crowd flows from available coarse-grained data sources, which can reduce the expense of urban systems \cite{liang2019urbanfm,ouyang2020fine}. Other tasks \cite{yao2018deep,zong2019deepdpm} are also actively studied by the community due to the vital impact of citywide crowd flow analytics.

Crowd flow analytics is not trivial as the traffic can be affected by multiple complex factors in spatio-temporal domains. As shown in Figure \ref{fig:intro}(a), the inflow of Region R1 is affected by outflows of nearby regions like R4 as well as distant regions, which indicates the spatial dependencies. For the temporal dependencies, crowd flow in a region is affected by recent, daily, and weekly historical traffic. To model the spatio-temporal dependencies, Convolutional Neural Networks (CNNs) have been widely used and achieved promising performance. A pioneering work \cite{zhang2016dnn} provided the first CNN-based method (DeepST) for modeling crowd flow, where convolution operators are used to extract spatially near and distant dependencies and the temporal dependencies are considered in different branches of networks. ST-ResNet \cite{zhang2017deep} further enhanced the performance of DeepST using residual structures. Very recently, a novel ConvPlus structure in DeepSTN+ \cite{lin2019deepstn} was proposed to learn the long-term spatial dependencies between two arbitrary regions. These CNN-based methods are characterized by two components: a complicated ST feature learner to capture features of the measurements, and a simple task-specific predictor to generate predictions on all regions. However, they have two main drawbacks:

\begin{figure}[!t]
  \centering
  \includegraphics[width=0.84\textwidth]{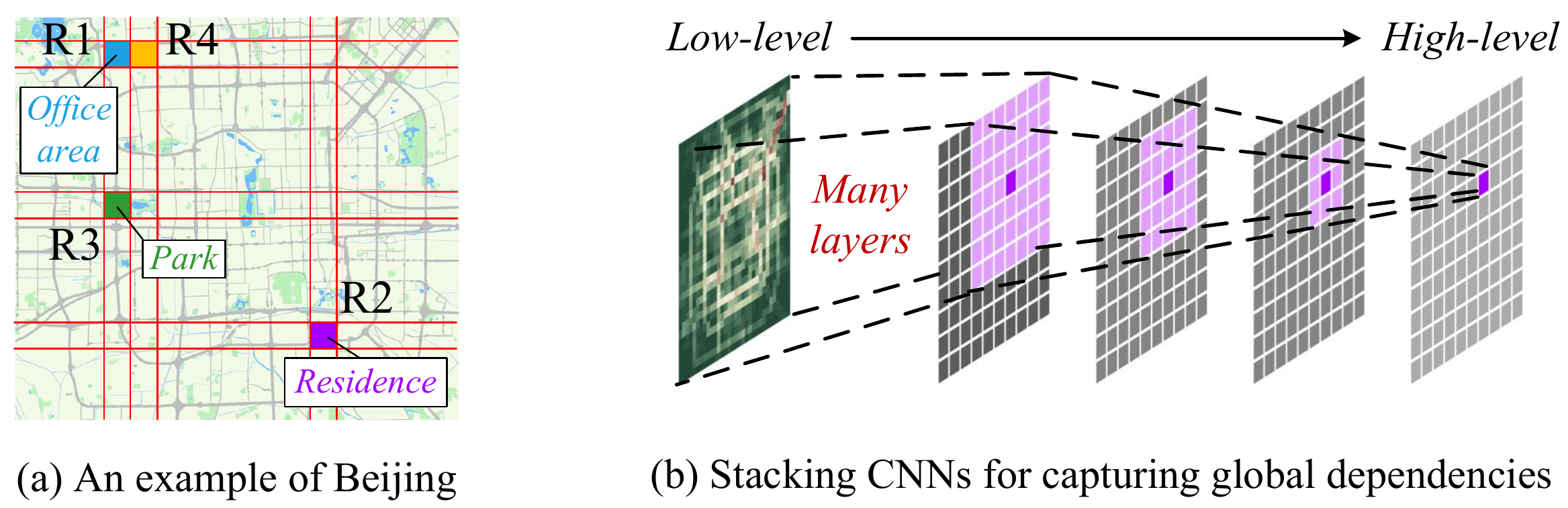}
  \caption{\label{fig:intro} Application of CNNs for citywide crowd flow analytics (Better view in color).}
\end{figure}
1) \emph{Inefficiency in learning global spatial dependencies}. Take traveling in Beijing (Figure \ref{fig:intro}) as an example. When predicting the inflow of R1 during morning hours, the outflow of distant regions like R2 needs to be considered, since it is common that people commute from a distant residence location. As people can travel around a modern city quickly, it becomes crucial to capture global spatial dependencies in this task. To this end, existing arts employ two approaches:
\begin{itemize}[leftmargin=*]
  \item \emph{Stacking CNNs to increase receptive fields}. Most previous studies like DeepST and ST-ResNet employ CNNs to capture information locally. But to capture global spatial dependencies, they have to stack many layers to increase the receptive field of the network (see Figure \ref{fig:intro}(b)). This is very inefficient since relationships between distant regions can only be captured by a near-top layer with a sufficiently large receptive field to cover all the regions of interest. 
  \item \emph{Learning long-range spatial dependencies directly}. 
  Instead of gradually increasing receptive fields, DeepSTN+ attempts to capture global spatial dependencies in \emph{every layer} using ConvPlus structure, which explicitly models all pairwise relationship between regions. However, a single layer of ConvPlus without pooling requires $O(n^2)$ parameters, where $n$ is the number of regions. Constrained by this bloated structure, DeepSTN+ cannot easily go deeper to learn higher-level representations for each region. Thus, how to learn global spatial dependencies more efficiently still remains a major challenge.
\end{itemize}

\begin{figure}[!t]
    \centering
    \includegraphics[width=0.75\textwidth]{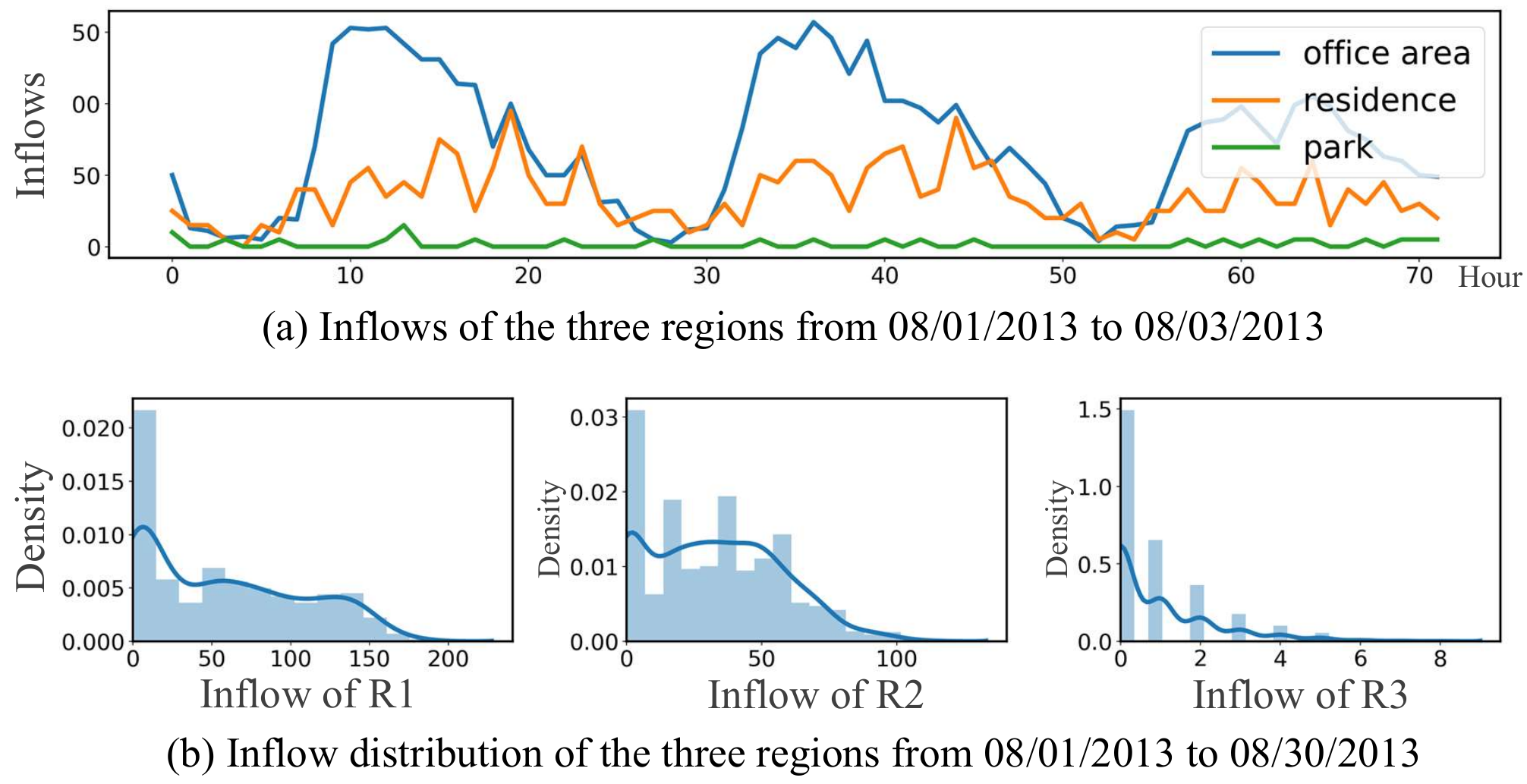}
    \caption{\label{fig:case} Illustration of daily patterns and inflow distribution in three regions.}
\end{figure}

2) \emph{Ignoring latent region functions}. Different from pixels in image processing, urban regions have different land functions according to their locations and surrounding POIs \cite{zheng2014urban,pan2019matrix}. Recall that R1, R2 and R3 in Figure \ref{fig:intro} correspond to an office area, a residential area and a park zone respectively. From Figure \ref{fig:case}(a), it can be seen easily that their daily patterns are entirely different. For instance, the office area (R1) usually reaches a traffic peak in the morning, while the residential area (R2) usually exhibits growth after dinner time. The difference between their daily flow distributions can also be seen from Figure \ref{fig:case}(b). However, the aforementioned methods have overlooked such varying latent functions among regions and used a simple predictor with shared parameters to predict flow for all regions, which inevitably resulted in degraded performance.
    
To address the above problems, we make the following contributions to the community. Primarily, we introduce DeepLGR, the first-ever general framework for raster-based crowd flow analytics. It is named according to how it stratifies a given task into three major procedures: 1)  \textbf{L}ocal feature extraction to learn representations for each region within small receptive fields; 2) {\textbf G}lobal context aggregation to efficiently capture the global spatial dependencies; and 3) {\textbf R}egion-specific prediction. Respectively,
\begin{itemize}[leftmargin=*]
    \item we present the first attempt to extract local region representations using Squeeze-and-Excitation networks (SENet)~\cite{hu2018squeeze}, which excels by including the channel-wise information as additional knowledge; 
    
    \item we design a global context module that firstly aggregates the region representations using a specific pooling method, and then upsample the global priors back to the original scale to generate global-aware features;
    
    \item we introduce a region-specific predictor based on tensor decomposition that factorizes the region-specific parameters of the predictor into a smaller core tensor and adjoint matrices. 
\end{itemize}

In addition, we evaluate our framework on two typical crowd flow analytics tasks: crowd flow forecasting \cite{zhang2016dnn,zhang2017deep} and fine-grained crowd flow inference \cite{liang2019urbanfm,ouyang2020fine}. Extensive experiments demonstrate the state-of-the-art performance and stability achieved by our framework. We have released our code at \url{https://github.com/yoshall/DeepLGR} for public use.

\section{Formulation}
In this section, we introduce several notations and formulate the problem of crowd flow analytics.  
As shown in Figure \ref{fig:intro}(a), we first follow the previous study \cite{zhang2016dnn} to partition an area of interest (e.g., a city) evenly into a $H \times W$ grid map based on longitude and latitude where a grid denotes a region. Thus, the crowd flow at a certain time $t$ can be denoted as a 3D tensor $\mathcal{P}_t \in \mathbb{R}^{H \times W \times K}$, where $K$ is the number of different flow measurements (e.g., inflow and outflow). Each entry $(i, j, k)$ denotes the value of the $k$-th measurement in the region $(i,j)$. 

Without loss of generality, we use $\mathcal{X} \in \mathbb{R}^{H \times W \times C}$ and $\mathcal{Y} \in \mathbb{R}^{H' \times W' \times D}$ as the input and output for a crowd flow analytics task, where $C$ and $D$ are the number of channels. For example, in the task of crowd flow prediction \cite{zhang2016dnn,zhang2017deep,lin2019deepstn}, the input is the historical observations $\mathcal{X}=\left\{\mathcal{P}_{i} | i=1,2, \cdots, \tau \right\} \in \mathbb{R}^{H \times W \times K\tau}$ and the target is to predict $\mathcal{Y}=\mathcal{P}_{\tau+1} \in \mathbb{R}^{H \times W \times K}$.

\section{Methodology}
\begin{figure}[!b]
    \centering
    \includegraphics[width=0.85\textwidth]{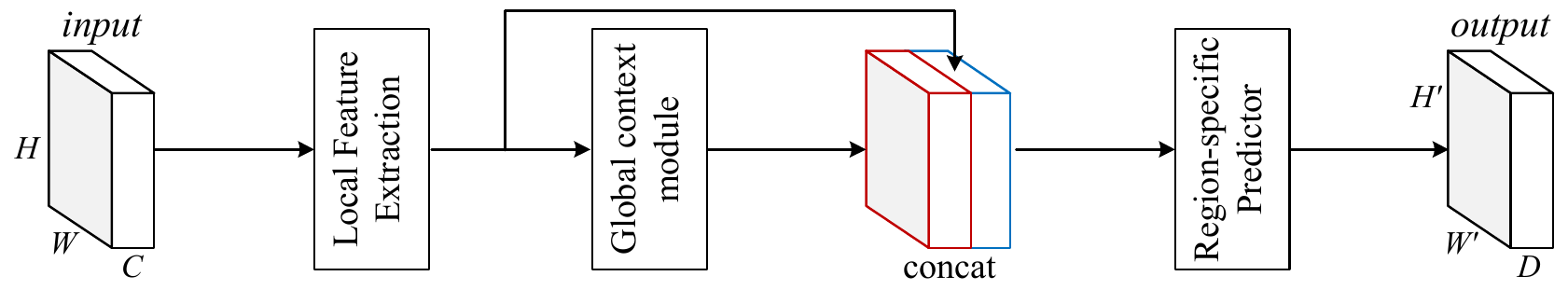}
    \caption{\label{fig:framework} The pipeline of DeepLGR, which contains three major components.}
\end{figure}

Figure \ref{fig:framework} presents the framework of DeepLGR, which can be easily generalized to all kinds of citywide crowd flow. Compared to the previous methods composed of an ST feature learner and a shared predictor for all regions, our framework contains three major components: local feature extraction, global context module and region-specific predictor. In the first component, we employ the SENet to learn representations for each region within small (i.e., local) receptive fields from the input tensor $\mathcal{X}$. To capture global spatial dependencies, we further design the global context module that considers the full region of interest. It first extracts global contextual priors from the learned region representations using a specific pooling method, and then upsamples the priors to the original scale to generate the global features. Once we obtain features from both local view and global view, we concatenate them into a tensor and then feed it to the region-specific predictor to make customized predictions for each region respectively.

For spatial dependencies, our framework employs the first two components which strategically capture both local-level (neighborhood) and global-level dependencies between regions. Following the mainstream CNN architectures for citywide crowd flow analytics \cite{zhang2016dnn,zhang2017deep,lin2019deepstn}, the temporal dependencies like closeness (recent), period (daily) and trend (weekly), if any, are considered in the channels of input. These temporal dependencies can interact with each other in the backbone network. Next, we will detail the three components respectively.

\subsection{Local Feature Extraction}
\begin{figure}[!h]
    \centering
    \includegraphics[width=0.6\textwidth]{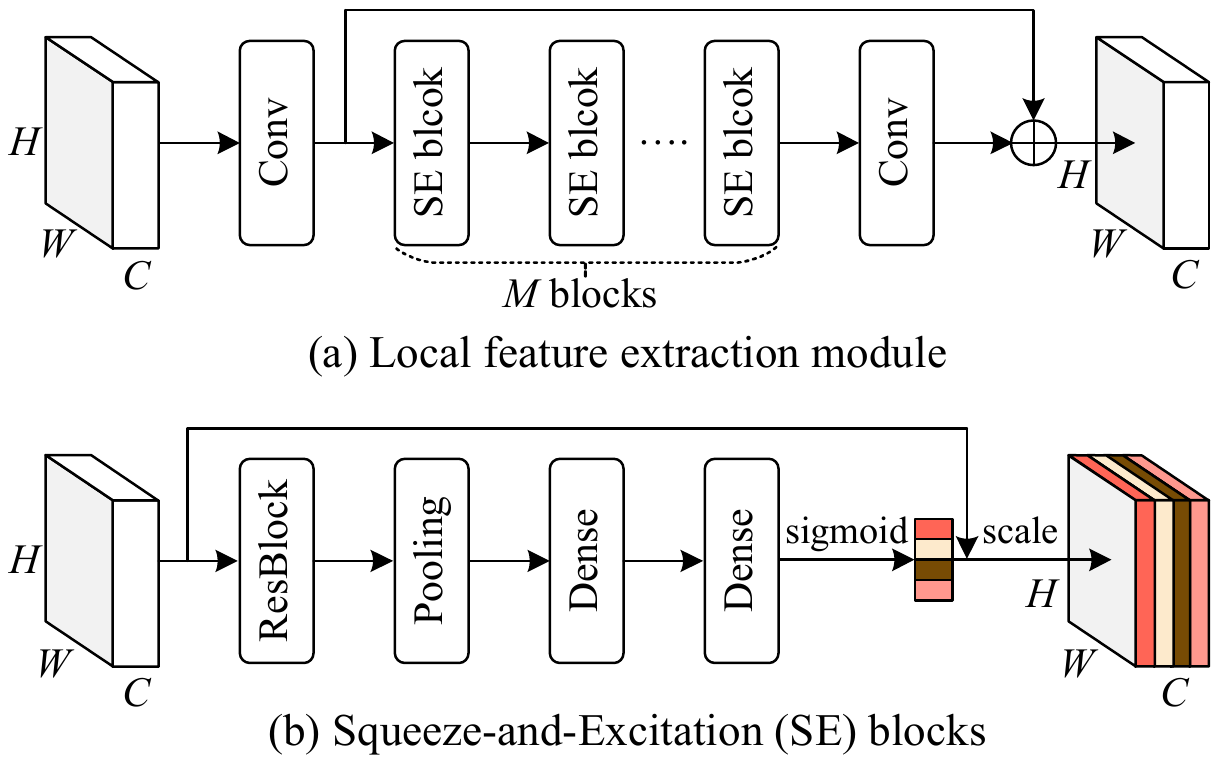}
    \caption{\label{fig:senet} The pipeline of local feature extraction, where the receptive fields depend on the number of SE blocks ($M$). Conv: convolutional layer. ResBlock: Residual block. Pooling: global average pooling. Dense: fully connected layer.}
\end{figure}

Recall that both the previous and current state-of-the-arts \cite{zhang2017deep,lin2019deepstn} use residual blocks to model the spatial dependencies from nearby regions. However, these methods mainly focus on the spatial dimension and have overlooked the channel-wise information in the feature maps. Thus, we employ SENet to fuse both spatial and channel-wise information within small (i.e., local) receptive fields at each layer, which has proven to be effective in producing compacted and discriminative features of each grid. Figure \ref{fig:senet}(a) illustrates the pipeline of the module for local feature extraction. The input is fed to a convolutional layer for initialization. Then, we stack $M$ squeeze-and-excitation (SE) blocks in Figure \ref{fig:senet}(b) for feature extraction, which is composed of three stages: 1) a residual block \cite{he2016deep} for feature learning; 2) a squeeze operation to squeeze global spatial information into a channel descriptor by global average pooling; 3) an excitation operation to fully capture the channel-wise dependencies: it first computes the attention coefficients over each channel via two fully connected layers followed by a sigmoid function, and then rescales the channels of original inputs by these weights. Finally, we use an output convolutional layer to transform the obtained high-level feature maps to the input of the next module. In summary, the SE structure enables this module to learn better representations for each region locally within receptive fields.

\subsection{Global Context Module}
\begin{figure}[!b]
    \centering
    \includegraphics[width=0.7\textwidth]{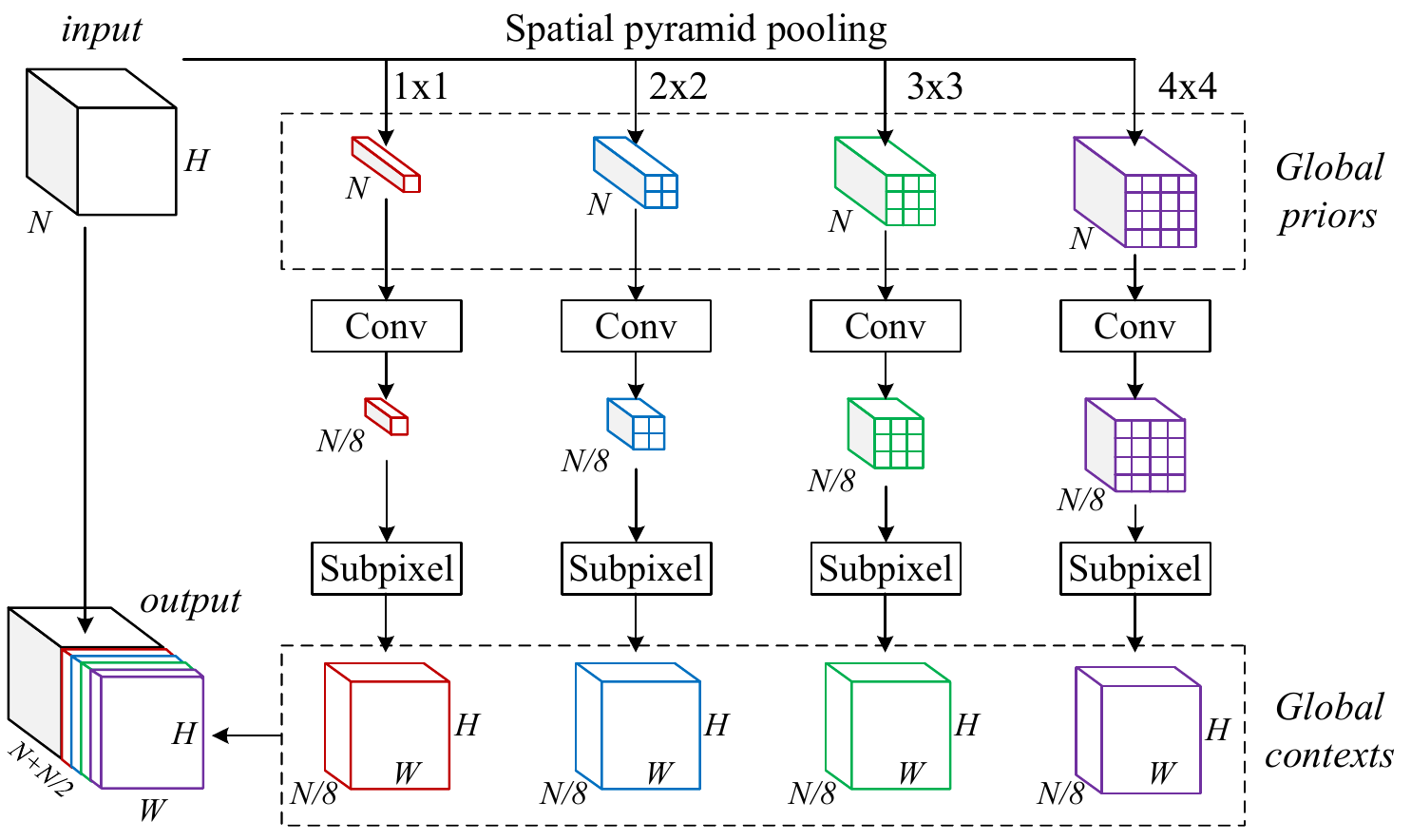}
    \caption{\label{fig:gcm} The pipeline of global context module, where Conv denotes a $1 \times 1$ convolutional layer for dimension reduction, and Subpixel contains a convolutional layer and a pixelshuffle operation sequentially to upsample the contextual priors. For simplicity, we use a 4-level pyramid (1$\times$1, 2$\times$2, 3$\times$3 and 4$\times$4) for an illustration.}
\end{figure}

After local feature extraction, we have designed a specific module that takes the output of the former component as input to generate global contexts for each region, so as to capture global spatial dependencies. As depicted in Figure \ref{fig:gcm}, we first employ spatial pyramid pooling \cite{he2015spatial} to generate a set of the global priors, where each prior is a spatially abstract of the original input under different pyramid scales. This operation allows the module to separate the feature map into different sub-regions and build pooled representation for different locations. For example, the 1$\times$1 prior (the red cube) denotes the coarsest level with only one single value at each channel, which is equivalent to global pooling operation that covers the whole image. In our experiments, we use a 4-level pyramid (1$\times$1, 2$\times$2, 4$\times$4 and 8$\times$8) to squeeze the input by average pooling. 

Once the global priors are obtained, an $1 \times 1$ convolution layer followed by a Batchnorm layer \cite{ioffe2015batch} is used for dimension reduction of channels from $N$ to $N/8$. Inspired by the study \cite{liang2019urbanfm} aiming at inferring fine-grained crowd flow from coarse-grained counterparts, we employ the Subpixel block \cite{shi2016real} to upsample the priors to generate new representations with the same size as the original inputs. For example, after the Subpixel block in 4$\times$4 branch, the output feature maps grow $H / 4$ and $W / 4$ times larger in height and width respectively with the number of channels unchanged. Different from PSPNet \cite{zhao2017pyramid} using bilinear interpolation for upsampling the priors, the Subpixel block considers the relationship between a super-region and its corresponding sub-regions by introducing a parametric design. Finally, we concatenate the input (i.e., region representations) with all levels of global features (i.e., context) as the output of this module. 

In summary, this module first converts the input feature map into priors (e.g., 1$\times$1 prior that encodes the information of all regions) and then upsamples the priors to learn the global-context-to-region influence (i.e., global spatial dependencies). Compared to the previous attempt (ConvPlus layer in DeepSTN+), our solution is more efficient and lightweight. Each ConvPlus layer directly models the pairwise relationships among all regions, thus demanding $O(n^2)$ parameters. With the increase of spatial granularity, it will induce extremely high computational costs due to the massive parameters. Thus, DeepSTN+ can hardly learn higher-level representations by simply increasing network depth. In contrast, as we have separated the procedures of local feature extraction and global context modeling, we can easily increase the network depth to gain better capacity.

\subsection{Region-Specific Predictor}
As mentioned before, each urban region has its unique land function. Previous studies \cite{zhang2016dnn,zhang2017deep,lin2019deepstn} mainly employ a single fully connected layer (equivalent to a $1\times1$ convolution) with shared weights as the predictor for all regions, which fails to capture this critical property. Thus, it is necessary to assign region-specific predictor to each region. 

\begin{figure}[!b]
    \centering
    \includegraphics[width=0.9\textwidth]{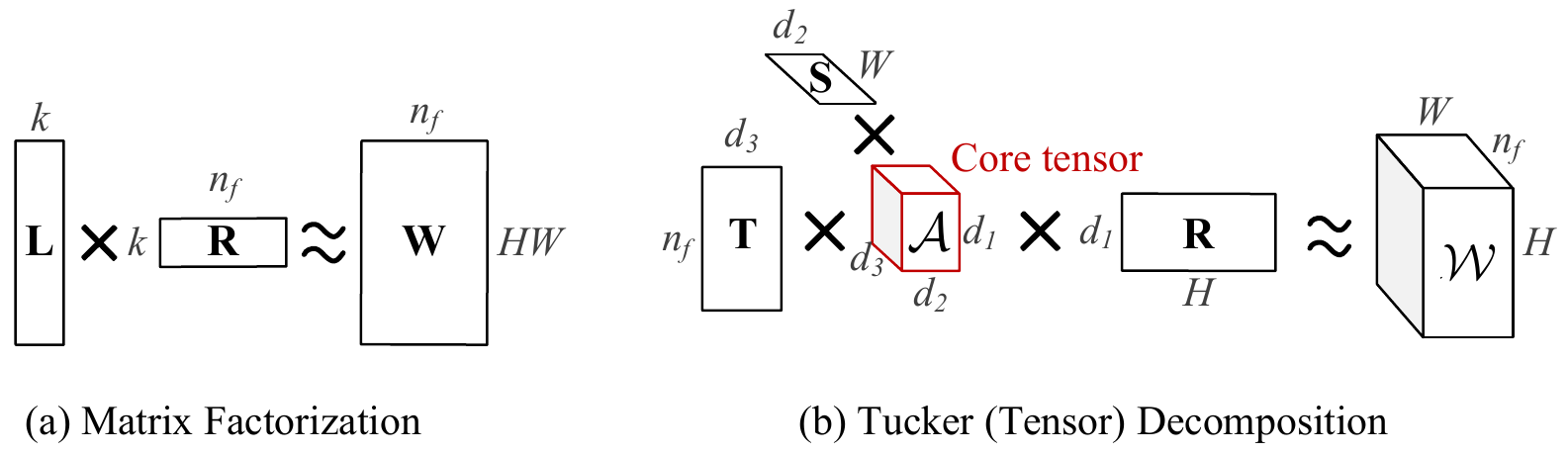}
    \caption{\label{fig:td} Illustration of matrix factorization and tensor decomposition.}
\end{figure}

Recall that the high-level feature obtained from last module is $\mathcal{Z} \in \mathbb{R}^{H \times W \times N'}$ and prediction result is $\mathcal{Y} \in \mathbb{R}^{H \times W \times D}$, where $N'=N+N/2$. Conventionally, the number of parameters in a shared fully connected layer is $n_f=N'D$. To achieve region-specific predictor, an intuitive solution is to use a customized fully connected layer for each region. However, it will induce $HW \times n_f$ parameters (denoted as a tensor $\mathcal{W} \in \mathbb{R}^{H \times W \times n_f}$), which can easily bloat up as the granularity increases. Recently, matrix factorization (MF) was used to avoid these drawbacks  \cite{pan2019matrix}, in which the parameter tensor $\mathcal{W}$ is reshaped to a matrix $\mathbf{W} \in \mathbb{R}^{HW \times n_f}$. As shown in Figure \ref{fig:td}(a), the authors from \cite{pan2019matrix} decompose the weight matrix $\mathbf{W}$ into two \emph{learnable} low-rank matrices, i.e., region embedding matrices $\mathbf{L} \in \mathbb{R}^{HW \times k}$ and parameter embedding matrices $\mathbf{R} \in \mathbb{R}^{k \times n_f}$. With the usage of MF, the number of the predictor parameters can be reduced to $(HW+n_f)k$, where $k \ll n_f$ and $k \ll HW$.

Nonetheless, directly flattening the parameter tensor $\mathcal{W}$ over the region dimension will lose the Euclidean structure of the flow map. For example, near things are more related than distant things according to the first law of geography, which indicates near regions should have similar prediction weights. Instead, we present a new idea for decomposing $\mathcal{W}$ using Tensor Decomposition (TD) \cite{tucker1966some}. It not only preserves the spatial similarity (dependencies) between regions, but also reduces the amount of parameters. As illustrated in Figure \ref{fig:td}(b), tensor $\mathcal{W}$ is decomposed into the multiplication of a core tensor $\mathcal{A} \in \mathbb{R}^{d_1 \times d_2 \times d_3}$ and three adjoint matrices, where $d_1$, $d_2$, and $d_3$ denote the number of latent factors for each matrix. The computation is as follows:
\begin{equation}
    \mathcal{W}=\mathcal{A} \times_{R} \mathbf{R} \times_{S} \mathbf{S} \times_{T} \mathbf{T},
\end{equation}
where $\times_{R}$ stands for the tensor-matrix multiplication; the subscript $R$ is the corresponding mode of the multiplication. For instance, $\mathbf{H}=\mathcal{A} \times_{R} \mathbf{R}$ is $\mathbf{H}_{i j k}=\sum_{i=1}^{d_1} A_{i j k} \times R_{i j}$. By this, we have changed the optimization target from $\mathcal{W}$ to the core tensor $\mathcal{A}$ as well as the three learnable matrices $\mathbf{R}$, $\mathbf{S}$ and $\mathbf{T}$. The core tensor is a low-rank representation summarising both the parametric and spatial information of the origin tensor $\mathcal{W}$. Compared to MF-based solution \cite{pan2019matrix}, our tensor decomposition can handle the higher-order relationships within the parameters. In addition, the number of parameters required is $d_1 d_2 d_3 + d_1 H + d_2 W + d_3 n_f$. Since $d_1$, $d_2$ and $d_3$ are usually very small, TD can achieve even much fewer parameters than MF, which is validated in our experiments.

\subsection{Optimization}
Since our framework is smooth and differentiable everywhere, it can be trained via the back-propagation algorithm. During the training phase, we use Adam optimizer to train our model by minimizing the entry-wise mean absolute error (MAE) between our prediction $\widehat{\mathcal{Y}}$ and the corresponding ground truth $\mathcal{Y}$:
\begin{equation}
    \mathcal{L}(\mathbf{\Theta})=\left\|\mathcal{Y}-\widehat{\mathcal{Y}}\right\|_1
\end{equation}
where $\mathbf{\Theta}$ denotes all learnable parameters in our framework.

\section{Experiments}\label{sec:exp}
To validate the generality of DeepLGR, we conduct experiments on two typical tasks of citywide crowd flow analytics: 
\begin{itemize}[leftmargin=*]
    \item \emph{Crowd flow forecasting}: This task is to forecast the inflow and outflow of each region in a city from historical readings. Following the settings of \cite{zhang2017deep}, we consider the temporal dependencies (i.e., closeness, period and trend) in different channels of input, and the output is the prediction of inflow and outflow for the next timestamp. Similar to \cite{zhang2017deep}, we set the length of closeness (recent), period (daily) and trend (weekly) to 5, 3, 3.
    
    \item \emph{Fine-grained flow inference}: In this task, we aim to infer fine-grained crowd flows throughout a city based on coarse-grained observations. We extend the state-of-the-art method named UrbanFM \cite{liang2019urbanfm} using our framework. Specifically, we replace the ResNet-based feature extraction of UrbanFM by our first component (SENet). Then, we add the global context module and region-specific predictor after the subpixel blocks in UrbanFM. 
\end{itemize}
\subsection{Experimental Settings}
\subsubsection{Datasets}
Two datasets were used in our experiments, including TaxiBJ and HappyValley. The former is the fine-grained version of the ones used by~\cite{zhang2017deep} and the latter is provided from~\cite{liang2019urbanfm}. Specifically, TaxiBJ consists of four different time spans (denoted as P1 to P4 with different number of taxicabs and distribution), while HappyValley is the hourly observations of human flow in a theme park in Beijing from ten months. The statistics are detailed in Table \ref{tab:dataset}. We select the flow data between 6am and 11pm to conduct our experiments. Using both datasets, we evaluate DeepLGR over the two aforementioned tasks:  In the first task, we employ the first 80\% data as training set, the next 10\% as validation set and the rest for test set;  In the second task, we follow all the experiment settings of \cite{liang2019urbanfm}, including training, validation and test set partition. The upscaling factors in TaxiBJ and HappyValley are 4 and 2 respectively.  
\begin{table}[!b]
  \caption{Dataset description.}
  \centering
  \small
  \tabcolsep=6.2mm
    \begin{tabular}{l|ll}
    \shline
    \textbf{Dataset} & \textbf{TaxiBJ} & \textbf{HappyValley} \\
    \hline
    Data type & Inflow and outflow & Staying flow \\
    Resolution & (128, 128) & (50,100) \\
    Sampling rate & 30 minutes & 1 hour \\
    \hline
      \  & P1: 07/01/2013-10/31/2013 &  \\
        Time Span    & P2: 02/01/2014-06/30/2014 & 01/01/2018- \\
        (mm/dd/yyyy)    & P3: 03/01/2015-06/30/2015 & 10/31/2018 \\
            & P4: 11/01/2015-03/31/2016 &  \\
    
    \shline
    \end{tabular}%
    \label{tab:dataset}
\end{table}%

\subsubsection{Evaluation Metrics}
We employ two widely-used criteria to evaluate our model from different aspects, including mean absolute error (MAE) and symmetric mean absolute percentage error (SMAPE). They are defined as:
\begin{equation*}
\text{MAE} = \frac{1}{z}\sum_{i=1}^{z}{ \left |{y_i-\tilde{y}_i} \right |, \:\: \text{SMAPE} = \frac{1}{z}\sum_{i=1}^{z}{\frac{\left |{y_i-\tilde{y}_i} \right|}{|y_i| + |\tilde{y}_i|}}},
\end{equation*}
where $y$ and $\tilde{y}$ are ground truth and predicted value respectively; $z$ is the total number of all entries. Smaller metric scores indicate better model performance.

\subsubsection{Baselines}
In the first task, we compare our framework with heuristics, time series methods and CNN-based baselines. Specifically, a naive method (\textbf{Last}) simply uses the last observation as the prediction result, and another heuristic (\textbf{CA}) leverages the closeness property to predict the future crowds by averaging the values from the previous 5 time steps. \textbf{ARIMA} is a well-known model for forecasting future values in a time series. Besides, the CNN-based baselines (including \textbf{DeepST} \cite{zhang2016dnn}, \textbf{ST-ResNet} \cite{zhang2017deep}, \textbf{ConvLSTM} \cite{shi2016real} and \textbf{DeepSTN+} \cite{lin2019deepstn}) have been introduced in Section \ref{para:intro}.

The second task was introduced only very recently by \cite{liang2019urbanfm}, where the authors presented the state-of-the-art method named \textbf{UrbanFM}. It considers the unique characteristics of this task, including the spatial hierarchy and external factors. Other strong baselines included in this work are related to image super-resolution, such as \textbf{VDSR} \cite{kim2016accurate} and \textbf{SRResNet} \cite{ledig2017photo}. We mainly use these three baselines for model comparison in this task. It is worth noting that all baselines are implemented with their default settings in both tasks. 

\subsubsection{Training Details \& Hyperparameters}
Our framework, as well as the above baselines, are fully implemented by Pytorch 1.1.0 with one GTX 2080TI. During the training phase, the learning rate is 0.005 and the batch size is 16. For the number of stacked SE blocks (denoted as $M$) in the first component, we conduct a grid search over $\{3,6,9,12\}$. For simplicity, we use the same hidden dimension (i.e., number of channels) at each 3$\times$3 convolutional layer in SE blocks, and conduct a grid search over $F=\{32,64,128\}$. 

\begin{table*}[!b]
  \caption{Prediction results on TaxiBJ over different time spans (P1-P4), where the bold number indicates the best performance of the column. We train and test each method five times, and present results using the format:``mean $\pm$ standard deviation".}
  \centering
  \footnotesize
  \tabcolsep=3mm
    \begin{tabular}{l|cc|cc}
    \shline
    \multicolumn{1}{l|}{\multirow{2}*{Method}} & \multicolumn{2}{c|}{P1} & \multicolumn{2}{c}{P2}  \\
\cline{2-5}        & MAE & SMAPE & MAE & SMAPE \\
    \hline

    CA  & 3.43  & 0.290  & 4.23  & 0.288 \\

    Last & 3.39  & 0.242  & 4.09  & 0.241  \\

    ARIMA & 3.08  & 0.403  & 3.53  & 0.385  \\

    DeepST &  2.59 $\pm$ 0.05  & 0.41 $\pm$ 0.01  & 2.94 $\pm$ 0.05  & 0.39 $\pm$ 0.01\\
    
    ST-ResNet & 2.53 $\pm$ 0.05  & 0.38 $\pm$ 0.05  & 2.93 $\pm$ 0.06  & 0.34 $\pm$ 0.07   \\
    
    ConvLSTM & 2.42 $\pm$ 0.02 & 0.41 $\pm$ 0.01  & 2.77 $\pm$ 0.01  & 0.39 $\pm$ 0.01   \\

    DeepSTN+ & 2.33 $\pm$ 0.04  & 0.35 $\pm$ 0.08  & 2.67 $\pm$ 0.02  & 0.32 $\pm$ 0.05  \\
    
    DeepLGR & \textbf{2.15 $\pm$ 0.00}  & \textbf{0.19 $\pm$ 0.00}  & \textbf{2.46 $\pm$ 0.00}  & \textbf{0.18 $\pm$ 0.00}  \\
    \shline
    \end{tabular}%
    
    \begin{tabular}{l|cc|cc}
    \shline
    \multicolumn{1}{l|}{\multirow{2}*{Method}} & \multicolumn{2}{c|}{P3} & \multicolumn{2}{c}{P4} \\
\cline{2-5}        & MAE & SMAPE & MAE & SMAPE\\
    \hline

    CA & 4.17  & 0.286 & 2.81  & 0.286  \\

    Last & 4.07  & 0.240  & 2.82  & 0.239  \\

    ARIMA & 3.68  & 0.363  & 2.61  & 0.420  \\

    DeepST & 2.97 $\pm$ 0.04  & 0.39 $\pm$ 0.01  & 2.16 $\pm$ 0.04  & 0.43 $\pm$ 0.02  \\
    
    ST-ResNet &  2.91 $\pm$ 0.06  & 0.33 $\pm$ 0.05  & 2.15 $\pm$ 0.04  & 0.32 $\pm$ 0.06  \\
    
    ConvLSTM & 2.87 $\pm$ 0.01  & 0.39 $\pm$ 0.01  & 2.09 $\pm$ 0.02  & 0.43 $\pm$ 0.02  \\

    DeepSTN+ & 2.82 $\pm$ 0.04  & 0.38 $\pm$ 0.05  & 2.05 $\pm$ 0.01  & 0.34 $\pm$ 0.05  \\
    
    DeepLGR & \textbf{2.56 $\pm$ 0.02}  & \textbf{0.19 $\pm$ 0.04}  & \textbf{1.84 $\pm$ 0.01}  & \textbf{0.19 $\pm$ 0.00}  \\
    \shline
    \end{tabular}%
  \label{tab:taxibj}%
\end{table*}%

\subsection{Results on Crowd Flow Forecasting}
\subsubsection{Model Comparison}
Here, we compare our framework with the baselines over the two datasets. We report the result of DeepLGR with $M=9$ and $F=64$ as our default setting. Further results regarding different $M$ will be discussed later.

Table \ref{tab:taxibj} shows the experimental results over P1 to P4 in TaxiBJ. We can observe that our framework clearly outperforms all baselines over both metrics. For instance, DeepLGR shows 10.2\% and 44.1\% improvements on MAE and SMAPE beyond the state-of-the-art method (DeepSTN+) in P4. The conventional model ARIMA performs much worse than deep learning models in these datasets, since it only considers the temporal dependencies among time series. Apart from the CNN-based methods, ConvLSTM advances DeepST and ST-ResNet because of the positive effect of its LSTM structure. However, it overlooks the global spatial dependencies between regions, which leads to inferiority compared to DeepSTN+ and DeepLGR. Another interesting observation is that the heuristics including CA and Last achieves much less SMAPE than previous CNN-based methods. Recall that SMAPE prefers to penalize the errors in regions with lower flow volumes. This observation reveals the importance of the temporal dependencies in such regions since CA and Last only consider the temporal closeness for forecasting. Only our method performs better than the heuristics on SMAPE with the usage of tensor decomposition, which will be detailed in the ablation study. Last but not least, DeepLGR is also more stable than the baselines according to the standard deviation observations.
\begin{table}[!b]
    \caption{Prediction results of various methods on the HappyValley dataset, where \#Params is the number of parameters and M denotes million.}
    \centering
    \tabcolsep=4.5mm
      \begin{tabular}{l|c|ccc}
      \shline
      Method & \#Params & MAE & SMAPE \\
      \hline
      CA  & x & 2.23  & 0.46   \\
      Last & x   & 2.20  & \textbf{0.38}\\
      ARIMA & 0.00M   & 2.14  & 0.47\\
      DeepST & 0.59M & 2.02 $\pm$ 0.05  & 0.56 $\pm$ 0.05 \\
      ST-ResNet & 2.73M & 1.98 $\pm$ 0.05  & 0.53 $\pm$ 0.04  \\
      ConvLSTM & 5.98M & 1.86 $\pm$ 0.01  & 0.48 $\pm$ 0.10   \\
      DeepSTN+ & 15.70M & 1.92 $\pm$ 0.01  & 0.54 $\pm$ 0.06 \\
      DeepLGR  & 0.97M & \textbf{1.84 $\pm$ 0.01}  & 0.40 $\pm$ 0.02   \\
  \shline    \end{tabular}%
    \label{tab:hv}%
  \end{table}%
  
Compared to TaxiBJ with a citywide scale, HappyValley focuses on a local area with a highly skewed flow distribution, where only a few regions contain dense populations. Table \ref{tab:hv} presents a comprehensive comparison of each model over this dataset. First, it can be seen easily that our framework shows great superiority against the CNN-based methods and slightly outperforms ConvLSTM in terms of both metrics, while using as little as 6.2\% of the amount of parameters required in the state-of-the-art method (DeepSTN+). This fact demonstrates that our model is more practical than other CNN-based solutions in real-world systems.
Second, similar to the results in TaxiBJ, DeepLGR performs more stable than the baselines according to the standard deviation in multiple experiments. 
Third, the heuristic method (Last) achieves the lowest SMAPE but the second-highest MAE, which can prove the skew distribution of this dataset. 
Last, the fact that DeepLGR and DeepSTN+ outperform ST-ResNet verifies the necessity of modeling global context in such a small area.

\subsubsection{Ablation Study}
To further investigate the effectiveness of each component, we compare DeepLGR with its variants over TaxiBJ-P1. For simplicity, we use the terms as local, global and TD to denote the three components in our framework respectively. Based on them, DeepLGR and its variants can be denoted as:
\begin{itemize} [leftmargin=*]
    \item \textbf{local+global+TD}: The original implementation of DeepLGR.
    \item \textbf{local+global+MF}: To show the effectiveness and lightweight property of TD against MF, we replace TD in the region-specific predictor by MF.
    \item \textbf{local+global}: Similar to the CNN-based baselines  \cite{zhang2016dnn,zhang2017deep,lin2019deepstn}, this variant uses shared parameters (i.e., not region-specific) as the predictor.
    \item \textbf{local+TD}: The variant of DeepLGR without global context module.
    \item \textbf{local+MF}: We first remove global context module from DeepLGR and then replace TD in region-specific predictor by MF.
    \item \textbf{local+bilinear}: We employ bilinear interpolation  rather than Subpixel block to upsample the global priors, so as to obtain new global representations. 
    \item \textbf{local}: The last two components are removed from DeepLGR. 
\end{itemize}

\begin{table}[!b]
      \caption{Results of different variants over TaxiBJ-P1 (trained/tested five times).}
      \centering
      \tabcolsep=4.5mm
        \begin{tabular}{l|c|cc}
        \shline
        Variants & \#Params & MAE & SMAPE \\
        \hline
        local & 0.72M   & 2.21 $\pm$ 0.01 & 0.37 $\pm$ 0.03 \\
        local+MF & 0.89M & 2.19 $\pm$ 0.02 & 0.36 $\pm$ 0.03 \\
        local+TD & 0.74M    & 2.19 $\pm$ 0.01 & 0.32 $\pm$ 0.03 \\
        \hline
        local+bilinear & 0.73M & 2.20 $\pm$ 0.02 & 0.35 $\pm$ 0.03 \\
        local+global & 2.30M   & 2.17 $\pm$ 0.02 & 0.29 $\pm$ 0.03 \\
        \hline
        local+global+MF & 2.46M    & 2.15 $\pm$ 0.00 & 0.27 $\pm$ 0.01 \\
        local+global+TD & 2.31M    & \textbf{2.15 $\pm$ 0.00}    & \textbf{0.19 $\pm$ 0.00} \\
        \shline
        \end{tabular}%
      \label{tab:variant}%
    \end{table}%

\begin{figure}[!t]
\centering
\includegraphics[width=0.8\textwidth]{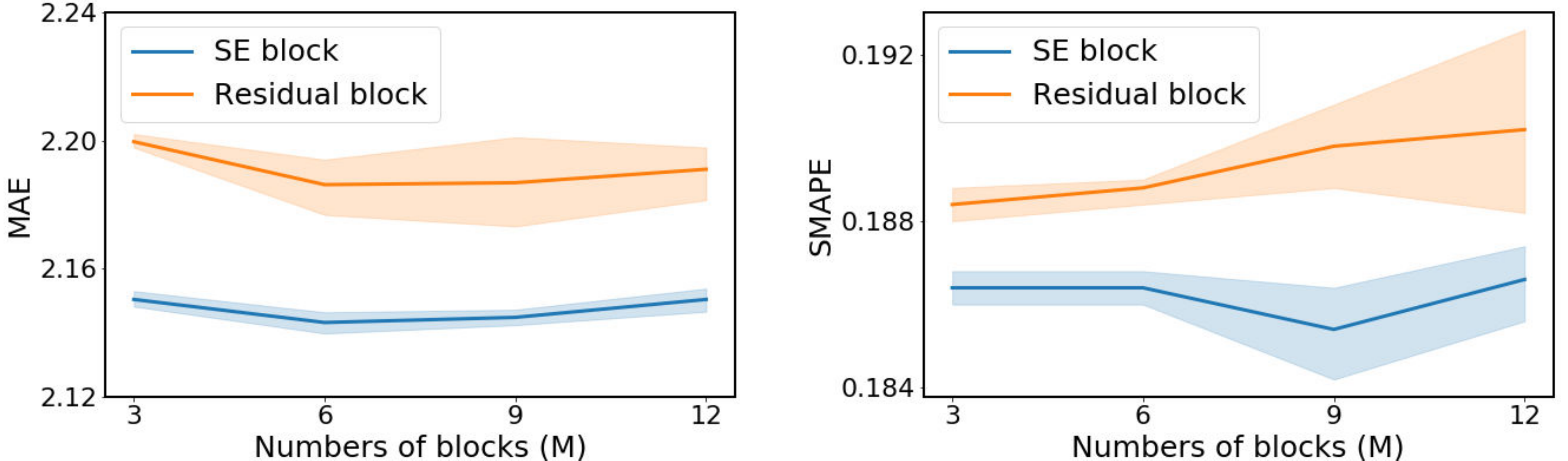}
\caption{\label{fig:effect_of_local} SE vs. residual block over P1, where the shade area is the standard deviation.}
\end{figure}

Table \ref{tab:variant} illustrates the variant comparison over TaxiBJ-P1. We discuss the effects of each model component as follows:
\begin{itemize}[leftmargin=*]
    \item \emph{Local feature extraction}: A powerful ST feature extractor enables the capability of extracting useful representations for each region. Compared to previous attempts like ST-ResNet based on residual blocks, our feature extraction module largely improves the performance (e.g., local vs. ST-ResNet in Table \ref{tab:taxibj} and \ref{tab:variant}).  We further investigate the effects of the number of SE blocks in this module. As shown in Figure \ref{fig:effect_of_local}, it achieves the best performance when $M=6$ in the test set. Noted that we choose $M=9$ as the default setting of DeepLGR because of its best performance on the validation set rather than the test set. Besides, we replace the SE blocks in this module by residual blocks to show the advantages of SE blocks, where the results are also in Figure \ref{fig:effect_of_local}.
    
    \item \emph{Global context module}: As a vital component in our framework, this module provides the global information to boost the performance. As illustrated in Table \ref{tab:variant}, the comparison between local and local+global (also local+TD and local+global+TD) can verify the effectiveness of this module. With the usage of Subpixel block with a parametric design, local+global brings an improvement beyond local+bilinear.
    
    \item \emph{Region-specific predictor}: This module is used to determine the region-specific parameters for predictions. Thus, we compare it with a shared fully connected layer with $n_f$ parameters (local+global), and the matrix decomposition method. From the last three rows of Table \ref{tab:variant}, we observe that TD demonstrates very competitive accuracy while using as little as 6.3 \% of the number of parameters required in MF (i.e., 0.01M vs. 0.16M). Moreover, TD significantly outperforms MF over SMAPE since it allows the model to capture spatial dependencies between regions.

\end{itemize}

\begin{table}[!b]
  \caption{Results of various models for fine-grained flow inference. We train/test each method five times, and present results using the format:``mean $\pm$ standard deviation".}
  \centering
  \tabcolsep=2.3mm
    \begin{tabular}{l|cc|cc}
    \shline
    \multirow{2}*{Method} & \multicolumn{2}{c|}{TaxiBJ-P1} & \multicolumn{2}{c}{HappyValley} \\
\cline{2-5}        & \multicolumn{1}{c}{MAE} & \multicolumn{1}{c|}{SMAPE} & \multicolumn{1}{c}{MAE} & \multicolumn{1}{c}{SMAPE} \\
    \hline
    VDSR & 2.23 $\pm$ 0.05  &  0.54 $\pm$ 0.03   & 2.13 $\pm$ 0.04     & 0.61 $\pm$ 0.02 \\
    
    SRResNet & 2.20 $\pm$ 0.05  & 0.52 $\pm$ 0.03  & 1.89 $\pm$ 0.05   & 0.61 $\pm$ 0.03  \\
    
    UrbanFM & 2.07 $\pm$ 0.03  & 0.25 $\pm$ 0.02  & 1.80 $\pm$ 0.02 & 0.41 $\pm$ 0.02  \\
    \hline
    local & 1.98 $\pm$ 0.01  & 0.20 $\pm$ 0.01  & 1.83 $\pm$ 0.01  & 0.43 $\pm$ 0.01  \\
    
    local+global & 1.96 $\pm$ 0.00  & 0.20 $\pm$ 0.01  & 1.78 $\pm$ 0.01  & 0.38 $\pm$ 0.01  \\
    
    local+global+TD & \textbf{1.95 $\pm$ 0.00}  & \textbf{0.18 $\pm$ 0.01}  & \textbf{1.76 $\pm$ 0.01}  & \textbf{0.35 $\pm$ 0.00}  \\
    \shline
    \end{tabular}%
  \label{tab:inference}%
\end{table}%

\subsection{Results on Fine-grained Flow Inference}
Experimental results on the second task have demonstrated the superiority of our framework again. From Table \ref{tab:inference}, we have the following observations: 
1) UrbanFM equipped with our framework (denoted as local+global+TD) shows considerable improvements against its original version on both datasets, validating its great generality in different applications. For example, DeepLGR achieves 5.8\% lower MAE and 28.0\% lower SMAPE than UrbanFM in the TaxiBJ-P1 dataset.
2) The three components of DeepLGR are effective according to the advancement of performance (only except local vs. UrbanFM in HappyValley).
3) Compared to VDSR and SRResNet for image-resolution , UrbanFM outperforms them by considering the domain knowledge, i.e., spatial hierarchy and external influence \cite{liang2019urbanfm}.
From above discussions, we can see that existing approaches like UrbanFM can be easily integrated with our framework.

\begin{figure}[!t]
  \centering
  \includegraphics[width=0.8\textwidth]{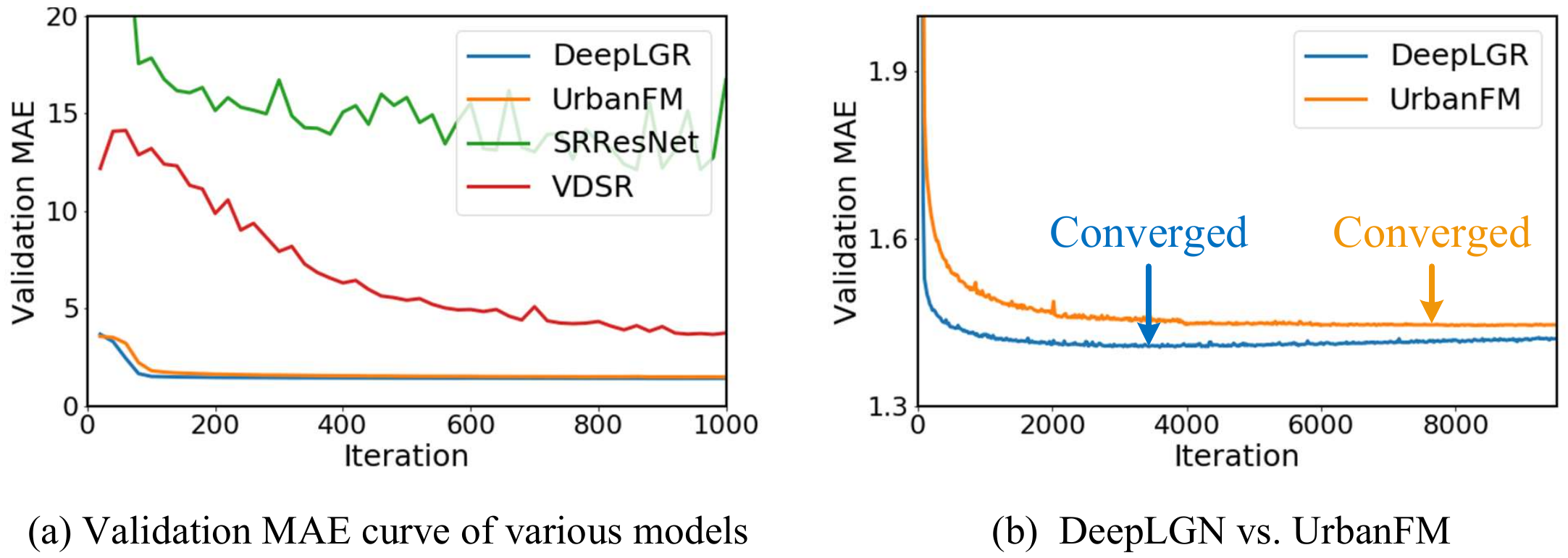}
  \caption{\label{fig:efficiency} Convergence speed of various methods over P1.}
\end{figure}

We further investigate the efficiency of DeepLGR. Figure \ref{fig:efficiency} plots the MAE on the validation set during the training phase using TaxiBJ-P1. Remarkably, UrbanFM and DeepLGR converge much smoother and faster than the others as shown in Figure \ref{fig:efficiency}(a). A more detailed comparison between UrbanFM and DeepLGR lies in Figure \ref{fig:efficiency}(b). From this figure, we can see that DeepLGR converges at iteration 3540 (epoch 37) while UrbanFM early-stops at iteration 7720 (epoch 81). This fact demonstrates that our framework can also accelerate the training phase of existing method.

\section{Related Work}
Citywide crowd flow analytics has attracted considerable attention of researchers in recent years. A series of studies have explored forecasting millions or even billions of individual mobility traces \cite{song2014prediction,fan2015citymomentum}. Different from analyzing crowd behaviors on an individual level, several works started to forecast citywide crowd flow by aggregating the crowds into corresponding regions \cite{li2015traffic,hoang2016fccf}. Among them, statistical learning was employed to capture inter-region relationship. With interest in obtaining fine-grained regional data, several studies \cite{liang2019urbanfm,zong2019deepdpm,ouyang2020fine} presented techniques to recover fine-grained crowd flow from coarse-grained data.

Recently, there have been many attempts focusing on end-to-end deep learning solutions such as CNNs for citywide crowd flow analytics. A pioneering study by \cite{zhang2016dnn} presented a general framework based on CNNs for citywide crowd flow prediction. By using a CNN architecture, their method can capture the spatio-temporal correlations reasonably and accurately. To overcome the gradient vanishing problem, they further integrated their framework using deep residual learning \cite{zhang2017deep}. Similar insight has been applied in taxi demand prediction \cite{yao2018deep}. Moreover, there are also several studies \cite{zonoozi2018periodic,yao2019revisiting} using RNNs to model the periodic temporal dependencies. Very recently, a ConvPlus structure \cite{lin2019deepstn} showed the state-of-the-art performance by directly modeling the long-range spatial dependencies between region pairs. However, as detailed in Section \ref{para:intro}, these methods are very inefficient in learning global spatial dependencies and none of them considers latent land function. To tackle these drawbacks, we have presented a general framework that can be easily generalized to all kinds of crowd flow data.

\section{Conclusion and Future Work}
In this paper, we have carefully investigated existing CNN-based methods for citywide crowd flow analytics, and exposed their inefficiency in capturing global spatial dependencies and incapability in generating region-specific predictions. Based on our discovery, we have presented the DeepLGR framework which decouples the local feature extraction and global context modeling, and provides a parameter-efficient solution for customizing regional outputs. We have evaluated DeepLGR over two real-world citywide crowd flow analytics tasks. In the prediction task, DeepLGR outperforms the state-of-the-art (DeepSTN+) by average 8.8\% and 45.9\% on TaxiBJ dataset, and 4.2\% and 25.9\% on HappyValley dataset in terms of MAE and SMAPE metrics respectively. Moreover, our framework is more lightweight than the state-of-the-art methods, which is very important in real practice. In the second task, we have verified that the existing approach can be easily integrated with our framework to boost its performance. In the future, we will explore two directions. First, we notice that manually designing neural networks requires amount of expert efforts and domain knowledge. To overcome this problem, we can follow a very recent study \cite{li2020autost} to study Neural Architecture Search (NAS), which can automatically construct a general neural network for diverse spatio-temporal tasks in cities. Second,  we will extend our framework to a much broader set of spatio-temporal tasks by using graph convolutions.  

\section*{Acknowledgement}
We thank all reviewers for their constructive and kind suggestions.
This work was supported by the National Key R\&D Program of China (2019YFB2101805) and Beijing Academy of Artificial Intelligence (BAAI).

%
%
%

\end{document}